\documentclass[envcountsame,oribibl]{llncs}
\usepackage{amssymb}
\usepackage{amsfonts}
\usepackage{amsmath}
\usepackage{dsfont} 
\usepackage{multirow}

\usepackage{graphicx}
\makeatletter
\newcommand{\figcaption}{\def\@captype{figure}\caption}
\newcommand{\tabcaption}{\def\@captype{table}\caption}
\makeatother

\usepackage[tight,footnotesize]{subfigure}

\usepackage[vlined,titlenumbered,algoruled]{algorithm2e}

\usepackage{colortbl}
\usepackage{ctable}
\newcolumntype{+}{>{\global\let\currentrowstyle\relax}}
\newcolumntype{^}{>{\currentrowstyle}}
\newcommand{\rowstyle}[1]{\gdef\currentrowstyle{#1}#1\ignorespaces}
\newcommand{\PreserveBackslash}[1]{\let\temp=\\#1\let\\=\temp}
\newcolumntype{C}[1]{>{\PreserveBackslash\centering}p{#1}}
\newcolumntype{R}[1]{>{\PreserveBackslash\raggedleft}p{#1}}
\newcolumntype{L}[1]{>{\PreserveBackslash\raggedright}p{#1}}

\usepackage{footmisc}
\DefineFNsymbols{mystyle}{\textdagger}
\setfnsymbol{mystyle}

\providecommand{\mb}[1]{\mathbf{#1}}

\providecommand{\mbx}{\mb{x}}
\providecommand{\mby}{\mb{y}}
\providecommand{\mbz}{\mb{z}}

\usepackage{mathtools}
\newtagform{myform}{\normalsize(}{)}
\usetagform{myform}

\begin{document}

\title{Stochastic Deep Compressive Sensing for the Reconstruction of Diffusion Tensor Cardiac MRI}

%
%

\author{Jo Schlemper\thanks{\scriptsize Co-first authors. Emails: \{jo.schlemper11,g.yang\}@imperial.ac.uk. J.S. has been funded by the EPSRC Programme Grant
(EP/P001009/1) and G.Y. has been funded by the British Heart Foundation (PG/16/78/32402). \textsuperscript{$\ddagger$}D.F. and D.R. are the co-last senior authors.}\inst{1}, Guang Yang\textsuperscript{$\dagger$}\inst{2}, Pedro Ferreira\inst{2}, Andrew Scott\inst{2}, Laura-Ann McGill\inst{2}, Zohya Khalique\inst{2}, Margarita Gorodezky\inst{2}, Malte Roehl\inst{2}, Jennifer Keegan\inst{2}, Dudley Pennell\inst{2}, David Firmin\textsuperscript{$\ddagger$}\inst{2}, \and Daniel Rueckert\textsuperscript{$\ddagger$}\inst{1}}
\institute{Department of Computing, Imperial College London, London, UK
\and National Heart \& Lung Institute, Imperial College London, London, UK and Cardiovascular Research Centre, Royal Brompton Hospital, London, UK}

\titlerunning{Deep Compressive Sensing for DT-CMR}


\maketitle \vspace{-2em}



\begin{abstract}
Understanding the structure of the heart at the microscopic scale of cardiomyocytes and their aggregates provides new insights into the mechanisms of heart disease and enables the investigation of effective therapeutics. Diffusion Tensor Cardiac Magnetic Resonance (DT-CMR) is a unique non-invasive technique that can resolve the microscopic structure, organisation, and integrity of the myocardium without the need for exogenous contrast agents. However, this technique suffers from relatively low signal-to-noise ratio (SNR) and frequent signal loss due to respiratory and cardiac motion. Current DT-CMR techniques rely on acquiring and averaging multiple signal acquisitions to improve the SNR. Moreover, in order to mitigate the influence of respiratory movement, patients are required to perform many breath holds which results in prolonged acquisition durations (e.g., $\sim$30 mins using the existing technology). In this study, we propose a novel cascaded Convolutional Neural Networks (CNN) based compressive sensing (CS) technique and explore its applicability to improve DT-CMR acquisitions. Our simulation based studies have achieved high reconstruction fidelity and good agreement between DT-CMR parameters obtained with the proposed reconstruction and fully sampled ground truth. When compared to other state-of-the-art methods, our proposed deep cascaded CNN method and its stochastic variation demonstrated significant improvements. To the best of our knowledge, this is the first study using deep CNN based CS for the DT-CMR reconstruction. In addition, with relatively straightforward modifications to the acquisition scheme, our method can easily be translated into a method for online, at-the-scanner reconstruction enabling the deployment of accelerated DT-CMR in various clinical applications.
%
%
\end{abstract}

\section{Introduction}
\vspace{-0.3cm}

Diffusion Tensor Cardiovascular Magnetic Resonance (DT-CMR) is a unique non-invasive technique, which provides rich structural and functional information on the myocardium at a microscopic scale, including parameters relating to the alignment and integrity of cardiomyocytes and aggregates of cardiomyocytes, known as sheetlets. Despite a long history of \emph{in vivo} DT-CMR and great efforts to drive the method towards a clinically usable technique \cite{Nielles-Vallespin_2013_AA}, its clinical use remains limited to research studies in a few specialist centres. These limited studies have produced a number of interesting findings in detection and diagnosis of ischemic heart disease, hypertrophic and dilated cardiomyopathies \cite{Ferreira_2014_AA,Deuster_2016_AA,Nielles-Vallespin_2017_AA}.

In comparison to the well established application of diffusion tensor imaging (DTI) in the brain, DT-CMR faces a number of additional challenges, including:  (1) the intrinsically low signal-to-noise ratio (SNR) of typical acquisition methods, which means that multiple signal acquisitions must be acquired and averaged to improve the net SNR; (2) signal loss caused by respiratory and cardiac motion during the application of diffusion sensitising gradients, means that these gradients must be short with respect to the motion and strong gradients must be used to provide sufficient sensitivity to the small diffusive movement of water molecules (on the order of microns) \cite{axel2017faster}; (3) the transverse relaxation time ($\mathrm{T_2}$) is substantially shorter in the myocardium than in the brain ($\sim$40 ms for the myocardium vs. $\sim$80 ms for the white matter of the brain). In contrast to neurological DTI, this dramatically limits the possible echo time (TE) that can be used in DT-CMR \cite{mekkaoui2017diffusion}; (4) the increased $\mathrm{B_0}$ inhomogeneity in the thorax may also result in more susceptibility-related distortions in the echo planar imaging (EPI) technique typically used for the DT-CMR \cite{mekkaoui2017diffusion}, which also limits spatial resolution. Some of these challenges have been partially addressed or have benefited from parallel imaging techniques for the in-plane acceleration, e.g., using SENSE (sensitivity encoding) and GRAPPA (generalized autocalibrating partially parallel acquisitions), or simultaneous multislice imaging \cite{lau2015accelerated}. However, typical acceleration factors are  normally limited to 2--3 \cite{mekkaoui2017diffusion}.

Compressive sensing (CS) is a promising technique for fast MRI \cite{lustig2007sparse} that circumvents the Nyquist-Shannon sampling criteria and can achieve a more aggressive acceleration. Comprehensive reviews of CS based fast MRI (CS-MRI) can be found elsewhere \cite{Hollingsworth2015}. Essentially, CS-MRI can obtain a \textit{perfect} reconstruction by using a \textit{nonlinear optimisation} on \textit{randomly undersampled} raw data, assuming the data or its transformation is \textit{compressible}. Although CS-MRI have been widely investigated \cite{Hollingsworth2015}, most previous studies have focused on the acceleration of the structural MRI, and only very few research studies have been conducted on DT-CMR \cite{wu2014accelerated,huang2016cardiac,ma2017accelerated}. These CS based fast DT-CMR methods demonstrated promising reconstruction results; however, the iterative nonlinear optimisation used in these methods requires a lengthy reconstruction procedure that could prevent their widespread usage, where there is a clinical need to view images immediately at the scanner. Therefore, a CS-based on-the-fly reconstruction of DT-CMR data would be highly desirable.

More recently, deep learning approaches have shown intriguing results in solving various medical image segmentation, registration and reconstruction problems. In particular, there are several deep learning based architectures that have been proposed for reconstruction of MRI data. The most widely used architecture is named U-Net \cite{ronneberger2015unet}, which is used to perform an end-to-end reconstruction \cite{lee2017deep,han2017deep}, and often combined with a residual learning \cite{lee2017deep, yang2017dagan} or generative adversarial networks \cite{yang2017dagan}. Alternative approaches have also been proposed, which modified the deep network architectures to embed traditional optimisation algorithms. These include gradient descent \cite{hammernik2017}, alternating direction method of multipliers (ADMM) \cite{NIPS2016_6406} or optimisation algorithms inspired by variable splitting techniques \cite{qin2017convolutional}. In addition, various clinical applications have been explored including knee imaging \cite{hammernik2017}, brain imaging \cite{yang2017dagan}, and dynamic cardiac imaging \cite{schlemper2017dynamic}.

In this study, a novel cascaded Convolutional Neural Networks (CNN) based CS technique has been proposed to simulate an efficient reconstruction of highly undersampled DT-CMR data. The proposed architecture improves upon the previously proposed networks by introducing dilated convolution instead of pooling to efficiently increase the receptive field. In addition, we introduce a novel \emph{stochastic} architecture, which is formulated by dropping the subnetworks at training. We show that this approach provides multifold benefits, including accelerated learning, improved robustness and an additional uncertainty estimate of the prediction. The following sections present the details of the proposed methodology, experimental set-up, achieved results and followed by discussions and conclusion.

\vspace{-0.3cm}

\section{Method}
\vspace{-0.3cm}

Let ${\bf{x}} \in \mathds{C}^N$ denote a complex-valued MR image to be reconstructed, represented as a vector with $N=N_xN_y$ where $N_x$ and $N_y$ are width and height of the image. Let ${\bf{y}} \in \mathds{C}^M$ $(M << N)$ represent the undersampled k-space measurements. Our problem is to reconstruct $\bf{x}$ from $\bf{y}$, formulated as an unconstrained optimisation:
\begin{equation}
  \label{eq:sparse_coding}
\begin{aligned}
& \underset{\mbx}{\text{argmin}} & & \mathcal{R}(\mbx) + \lambda \| \mby - \mb{F}_u \mbx \|^2_2
\end{aligned}
\end{equation}
Here $\mb{F}_u$ is an undersampling Fourier encoding matrix, $\mathcal{R}$ expresses regularisation terms on $\mbx$ and $\lambda$ is a hyper-parameter often associated to the noise level. 

~\\
\noindent\textbf{Deep Cascaded CNN~}In general, the regularisation terms $\mathcal{R}$ in Eq. \ref{eq:sparse_coding} can be non-convex (such as $\ell_0$ in the sparsifying domain). Therefore, traditionally, one introduces an auxiliary variable $\bf{z}$ as variable splitting technique and solves the following penalty functional:

\begin{equation}
\label{eq: penalty_function}
\underset{\mbx,\mbz}{\text{argmin }} \mathcal{R}(\mbz) + \lambda \| \mby - \mb{F}_u \mbx \|_2^2 + \mu \|\mbx-\mbz\|^2_2
\end{equation}
where $\mu$ is a penalty parameter. By applying alternating minimisation over $\bf{x}$ and $\bf{z}$, Eq. \ref{eq: penalty_function} can be solved via the following iterative procedures:
\begin{subequations}
\label{eq:alternate_minimisation} 
\begin{align}
\mbz^{(i)} & = \underset{\mbz}{\text{argmin }}   \mathcal{R}(\mbz) + \mu \| \mbx^{(i)} - \mbz \|^2_2  \label{eq:proximal_operator}
\\ 
\mbx^{(i+1)} & = \underset{\mbx}{\text{argmin }}  \lambda \| \mby - \mb{F}_u \mbx \|^2_2 + \mu \| \mbx - \mbz^{(i)} \|^2_2 \label{eq:data_fidelity}
\end{align}
\end{subequations}
where $\mbx^{(0)} = \mbx_u = {\bf{F}}_u^H y$ is the zero-filled (ZF) reconstruction taken as an initialisation and $\mbz$ can be seen as an intermediate state of the optimisation process. For MRI reconstruction, Eq. \ref{eq:data_fidelity} is often regarded as a \emph{data consistency} (DC) step where we could obtain the following closed-form solution \cite{schlemper2017deep}:

\begin{small}
\begin{equation}
\begin{array}{l}
\mbx^{(i+1)} = \textnormal{DC}(\mbz^{(i)};\mby, \lambda_0, \Omega) = \mb{F}^{H}\mb{\Lambda}\mb{F}\mbz^{(i)} + \frac{\lambda_0}{1 + \lambda_0} \mb{F}_u^{H}\mby, 

\mb{\Lambda}_{kk} =
\begin{cases}
  1 & \text{if } k \not \in \Omega \\
  \frac{1}{1+\lambda_0} & \text{if } k \in \Omega
\end{cases}
\label{eq:dc_fnc}
\end{array}
\end{equation}
\end{small}which $\mb{F}$ is the full Fourier encoding matrix (a discrete Fourier transform in this case), $\lambda_0 = \lambda/\mu$ is a ratio of regularization parameters from Eq. \ref{eq:dc_fnc}, $\Omega$ is an index set of the acquired $k$-space samples and $\mb{\Lambda}$ is a diagonal matrix. Eq. \ref{eq:proximal_operator} is the proximal operator of the prior $\mathcal{R}$, and instead of explicitly determining the form of the regularisation term, one can learn the proximal operator by using the CNN directly. In so doing,  iterative reconstruction with a cascaded CNN and DC steps is performed. The whole framework can be optimized end-to-end, yielding one cascaded deep network. This network is referred to as \emph{DC-CNN}. 

~\\ 
\noindent\textbf{Stochastic DC-CNN~} Inspired by \cite{stochastic_depth}, we extend the DC-CNN framework into a \emph{stochastic DC-CNN} (s-DC-CNN)---during the training, the $i$-th subnetwork is dropped with a probability of $p = (i-1) / 2n_c$, where $n_c$ is the total number of the cascaded CNN. During the testing (inference), we can use all of the subnetworks to perform the reconstruction, which is expected to provide the best performance as the most depths are used. Alternatively, we could sample the network configurations $\theta$ using the distribution $P$ from the above strategy, and then reconstruct the image as an ensemble of the sampled model $\bar{x} = \mathbb{E}_{\theta \sim P}(f(x_u|\theta))$. In so doing, the variance of the predicted value can be used as an \emph{uncertainty estimate} given the ensemble. This alternative approach provides several benefits: (1) it accelerates the learning because the expected depth of the network is much shorter; (2) this simultaneously helps the error terms to be backpropagated better as the depth is shortened, allowing us to train deeper networks if desired; (3) due to the stochastic connection, each subnetwork can see different levels of residual noise that help the network learn better and become more robust.

In addition, we used dilated convolution to efficiently increase the receptive field. In the original DC-CNN framework, each denoising subnetwork only consists of 5 layers of 3$\times$3 2D convolution layers, which has a limited receptive field of size 11.  In our new s-DC-CNN framework, we employed the dilated convolution except for the first and the last convolution that has a receptive field of size 23. It is of note that compared to pooling operation, the dilated convolution can avoid the needs of upsampling, and subsequently prevents information loss and interpolation artefacts, and also keeps the network more compact. Furthermore, we employed batch-normalisation to improve the training and applied leaky rectified linear unit with $\alpha=0.01$. 

\vspace{-0.3cm}

\section{Results}
\vspace{-0.3cm}

\noindent\textbf{Experimental Settings~} In order to test the efficacy of our proposed DC-CNN method for the DT-CMR reconstruction, we performed the following simulation based studies. First, we simulated undersampled DT-CMR datasets using \emph{in vivo} DT-CMR data acquired at peak-systole or in diastasis in healthy volunteers. The data were acquired using a breath hold stimulated echo acquisition mode EPI sequence \cite{Nielles-Vallespin_2013_AA}, with diffusion encoded over 1 complete cardiac cycle (detailed scanning parameters in Supplementary Materials). Evaluation has been carried out using 178 DT-CMR scans.

The image data were firstly converted into k-space data using an inverse Fourier transform and undersampled by a pseudo-random 1D undersampling mask. Each line was sampled from a Gaussian distribution with mean centred at the origin of $k$-space and variance proportional to the extent of $k$-space. We applied various random undersampling patterns to the data acquired for different diffusion weighting directions and different signal averages in order to make the trained network generalise better on various aliasing artefacts. Note that in this work, the magnitude images were used. However, the effectiveness of the method to complex-valued data has been previously verified in \cite{schlemper2017deep}.

The studied scans were split into independent training, validation and test sets containing 142, 17 and 19 cases, respectively. 
In the training stage, both information from fully-sampled data and undersampled data were used. In the testing (inference) stage, the zero-filled undersampled data were input to the trained network to yield a reconstruction.
~\\

\noindent\textbf{Comparison Studies~} To compare with other state-of-the-art conventional CS-MRI and recently proposed deep learning based methods, we re-implemented the following methods for the DT-CMR reconstruction, including a dictionary learning based CS-MRI (i.e., \emph{DLMRI} \cite{ravishankar2011mr}) and a U-Net based method (\emph{UNET-CS}). For the proposed approach, we selected to use 6-layer subnetwork with $n_c= 15$ that yields a 105-layer network including the DC layers. Note that the hyper-parameters were chosen from a grid search at a coarse scale but is by no means optimal. We performed an ablation study to test the benefits of the proposed improvements. The original network without and with dilated convolution are denoted as \emph{DC-CNN} \cite{schlemper2017dynamic} and \emph{DC-CNN-d}. The proposed stochastic version without and with dilated convolution are referred as \emph{s-DC-CNN} and \emph{s-DC-CNN-d} (the detailed network architectures are described in the supplementary materials). For the stochastic networks, we used all subnetworks for reconstruction except for when generating the uncertainty map. 
~\\

\noindent\textbf{Training Settings~} Each network was trained for 100 epochs using  mean squared error loss between the network reconstruction and the ground truth. In the first 50 epochs, the network was trained on various undersampling factors (UF) of (0--12$\times$). For the last 50 epochs, the network was fine-tuned for the UF of \{2$\times$, 5$\times$, 8$\times$\} individually for each experiment. We used the Adam optimiser with a learning rate of $10^{-4}$, that was subsequently reduced by a factor of 10 for every 20 epochs during the fine-tuning. 

\begin{table}[!htbp]
  \setlength{\abovecaptionskip}{0pt} 
  \setlength{\belowcaptionskip}{0pt} 
  \caption{\scriptsize{Quantitative results [mean (std)] of the 19 independent testing subjects.}}
  \centering
  \scalebox{.65}{
  \setlength{\floatsep}{10pt plus 3pt minus 2pt} 
\begin{tabular}{p{0.8cm}^p{2cm}^p{2.4cm}^p{2.2cm}^p{2.2cm}^p{2.2cm}^p{2cm}^p{2cm}^p{2cm}}
\addlinespace
\toprule\rowstyle{}
 \bfseries{UF} & \bfseries{Model}       & \bfseries{s-DC-CNN-d} & \bfseries{s-DC-CNN} & \bfseries{DC-CNN-d} & \bfseries{DC-CNN} & \bfseries{UNET-CS} & \bfseries{DLMRI} & \bfseries{Zero-Filled} \\ \midrule
2$\times$& \bfseries{PSNR} & \bfseries{37.747 (2.41)}& 37.667 (2.35)& 37.64 (2.33)& 37.73 (2.36) & 33.16 (2.14)& 33.62 (2.26) & 27.89 (2.08) \\
5$\times$& \bfseries{PSNR} & \bfseries{30.96 (2.10)} & 30.85 (2.13) & 30.88 (2.08)& 30.8 (2.1)   & 28.40 (2.44)& 28.40  (2.44) & 23.71 (1.72) \\
8$\times$& \bfseries{PSNR} & \bfseries{28.81 (2.00)} & 28.50 (2.01) & 28.79 (1.97)& 28.62 (2.00) & 27.35 (2.00)& 25.88 (2.36) & 22.71 (1.64) \\ \midrule
 \multirow{3}{*}{5$\times$} 
& \bfseries{FA RMSE}    &      \bfseries{  0.08 (0.03)}&      \bfseries{0.08(0.03)} &      0.09(0.03) &    0.09(0.03) &   0.11(0.05) &   0.11(0.04) &   0.18(0.04) \\
& \bfseries{MD RMSE}    &     \bfseries{ 0.11(0.05)}&      0.12(0.05) &      \bfseries{0.11(0.05)} &    0.12(0.05) &   0.16(0.08) &   0.15(0.07) &   0.19(0.10) \\
& \bfseries{HA RMSE}    &     \bfseries{13.34(2.93)}&     13.97(3.11) &     13.68(2.91) &   14.04(3.04) &  16.29(3.27) &  17.88(2.9) &  26.87(3.31) \\
\bottomrule
\end{tabular}
  }
  \label{table:psnr}
\end{table}

\noindent{\textbf{Results~}} Table \ref{table:psnr} tabulates mean and standard deviation (std) of the peak signal-to-noise ratio (PSNR) we obtained by various CS-MRI methods. For the three different UF we tested, the proposed s-DC-CNN-d obtained the best performance, and all our DC-CNN variations outperformed the DLMRI and the UNET-CS. This has been further confirmed by the root mean-squared-error (RMSE) calculated from the computed fractional anisotropy (FA), mean diffusivity (MD) and helix angle (HA) of the 5$\times$ undersampling cases.

\begin{figure}[tb]
\centering
\setlength{\abovecaptionskip}{0pt} 
\setlength{\belowcaptionskip}{0pt}
\scalebox{1}{\includegraphics[width=1\textwidth]{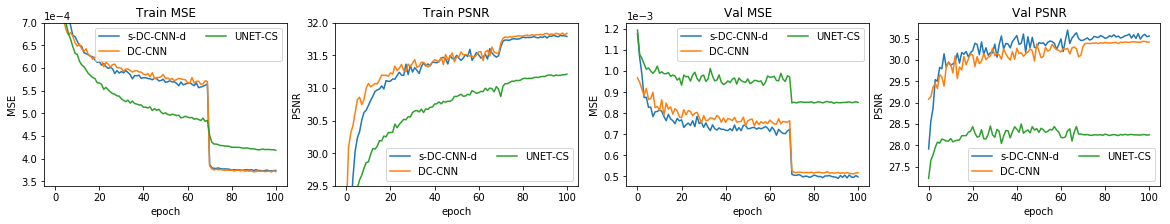}}
\figcaption{\scriptsize{Convergence analysis of the s-DC-CNN-d compared to the DC-CNN and the UNET-CS.}}
\label{fig:train}
\end{figure}

Convergence analysis (Figure \ref{fig:train}) shows that the proposed s-DC-CNN-d learned slower, but eventually generalised better. The fact that UNET-CS was quickly overfitted may be due to low SNR in the original images that hamper the network to learn a meaningful end-to-end mapping. Compared to the DC-CNN, the proposed s-DC-CNN-d obtained a better PSNR with less epochs during the validation that can be attributed to the fact that the proposed s-DC-CNN-d  has an accelerated learning.

\begin{figure}[tb]
\centering
\setlength{\abovecaptionskip}{0pt} 
\setlength{\belowcaptionskip}{0pt}
\scalebox{.95}{\includegraphics[width=1\textwidth]{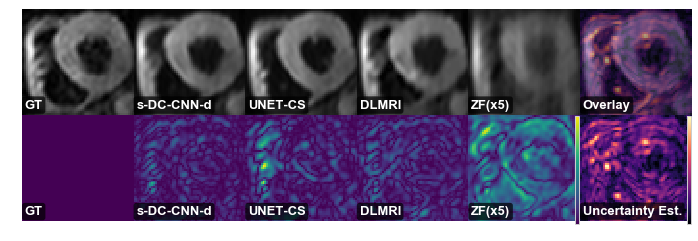}}
\figcaption{\scriptsize{Qualitative comparison of the reconstructions and error maps for the 5$\times$ undersampling.}}
\label{fig:error_map}
\end{figure}

Qualitative visualisation (Figure \ref{fig:error_map}) demonstrates that perceptually both DLMRI and the UNET-CS are over-smoothed in their reconstructed results with clearly larger errors, and the textural details were better preserved in the results obtained by the proposed s-DC-CNN-d. From a qualitative analysis, we found that the generated uncertainty estimate using our s-DC-CNN-d highlighted the most challenging area for our algorithms to reconstruct, e.g., the edges of the ventricle and the highly blurred areas in the undersampled input. 


\begin{figure}[tb]
\centering
\setlength{\abovecaptionskip}{0pt} 
\setlength{\belowcaptionskip}{0pt}
\scalebox{1}{\includegraphics[width=1\textwidth]{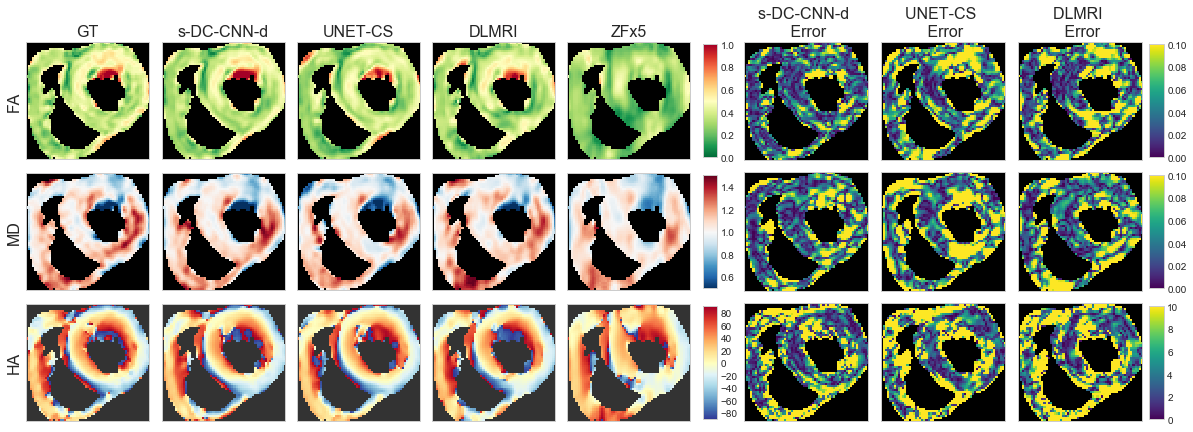}}
\figcaption{\scriptsize{Comparison of the diffusion tensor parameters and error maps. From top to bottom: fractional anisotropy (FA), mean diffusivity (MD) ($10^{-3} \textrm{mm}^2 \textrm{s}^{-1}$) and Helix-angle (HA) (degrees).}}
\label{fig:dti_error_map}
\end{figure}

Figure \ref{fig:dti_error_map} shows the computed FA, MD and HA for the 5$\times$ undersampling. Overall, our proposed s-DC-CNN-d achieved better calculated diffusion tensor parameters compared to the ones obtained by other methods. In particular, results obtained by the s-DC-CNN-d have a more smooth transition of the HA from epicardial to endocardial surface in the
normal left ventricular free wall that resembles the HA calculated by the fully sampled ground truth data. 

Finally, we timed the reconstruction of each algorithm (on a GeForce GTX 1080 GPU) and obtained (1) s-DC-CNN-d: 0.065$\pm$0.03 s per frame, (2) s-DC-CNN: 0.04$\pm$0.02 s per frame, (3) DC-CNN-d: 0.052$\pm$0.02 s per frame, (4) DC-CNN: 0.04$\pm$0.02 s per frame and (5) UNET-CS: 0.003$\pm$0.04 s per frame. The DLMRI method has only a CPU implementation and takes $\approx 60$s per iteration, and we used 400 iterations per image ($>$6 hours). 

\vspace{-0.3cm}

\section{Discussion and Conclusion}
\vspace{-0.3cm}

In this study, we proposed a novel deep cascaded CNN based CS-MRI and its stochastic variation for the DT-CMR reconstruction. To the best of our knowledge, it is the first work to consider such application to DT-CMR although the current study is nevertheless simulation based. In addition, we are the first to consider such a stochastic formulation to take the variance of the reconstruction models into account that can be visualised via an uncertainty map. Compared to other state-of-the-art methods using dictionary learning or U-Net based deep learning architecture, our proposed method incorporated dilated convolution and achieved significantly superior reconstruction fidelity with very efficient computation that can be translated into a real-time reconstruction scheme working directly on the scanner. As a future direction, we will carry on further development from on our current simulation based study to accelerate DT-CMR reconstruction and improve its spatial resolution and we can envisage its deployment in various clinical applications.

\vspace{-0.3cm}




\bibliographystyle{splncs03}
\bibliography{ref}

\begin{thebibliography}{10}
\providecommand{\url}[1]{\texttt{#1}}
\providecommand{\urlprefix}{URL }

\bibitem{axel2017faster}
Axel, L.: Faster diffusion-weighted {MR} imaging of cardiac microstructure.
  Radiology  282(3),  622--626 (2017)

\bibitem{Deuster_2016_AA}
von Deuster, C., et~al.: Studying dynamic myofiber aggregate reorientation in
  dilated cardiomyopathy using in vivo magnetic resonance diffusion tensor
  imaging. Circ. Cardiovasc. Imag.  9(10) (2016)

\bibitem{Ferreira_2014_AA}
Ferreira, P.F., et~al.: In vivo cardiovascular magnetic resonance diffusion
  tensor imaging shows evidence of abnormal myocardial laminar orientations and
  mobility in hypertrophic cardiomyopathy. J. Cardiovasc. Magn. Reson.  16, ~87
  (2014)

\bibitem{hammernik2017}
Hammernik, K., et~al.: Learning a variational network for reconstruction of
  accelerated {MRI} data. arXiv preprint arXiv:1704.00447  (2017)

\bibitem{han2017deep}
Han, Y., et~al.: Deep learning with domain adaptation for accelerated
  projection-reconstruction {MR}. Magn. Reson. Med.  (2018)

\bibitem{Hollingsworth2015}
Hollingsworth, G.: {Reducing acquisition time in clinical MRI by data
  undersampling and compressed sensing reconstruction}. Phys. Med. Biol.
  60(21),  297--322 (2015)

\bibitem{stochastic_depth}
Huang, G., et~al.: Deep networks with stochastic depth. arXiv preprint
  arXiv:1603.09382  (2017)

\bibitem{huang2016cardiac}
Huang, J., et~al.: Cardiac diffusion tensor imaging based on compressed sensing
  using joint sparsity and low-rank approximation. Technology and Health Care
  24(s2),  S593--S599 (2016)

\bibitem{lau2015accelerated}
Lau, A.Z., et~al.: Accelerated human cardiac diffusion tensor imaging using
  simultaneous multislice imaging. Magn. Reson. Med.  73(3),  995--1004 (2015)

\bibitem{lee2017deep}
Lee, D., et~al.: Deep residual learning for compressed sensing {MRI}. In:
  International Symposium on Biomedical Imaging. pp. 15--18. IEEE (2017)

\bibitem{lustig2007sparse}
Lustig, M., et~al.: Sparse {MRI}: The application of compressed sensing for
  rapid {MR} imaging. Magn. Reson. Med.  58(6),  1182--1195 (2007)

\bibitem{ma2017accelerated}
Ma, S., et~al.: Accelerated cardiac diffusion tensor imaging using joint
  low-rank and sparsity constraints. IEEE Trans. on Biomed. Eng.  (2017)

\bibitem{mekkaoui2017diffusion}
Mekkaoui, C., et~al.: Diffusion {MRI} in the heart. NMR in Biomedicine  30(3)
  (2017)

\bibitem{Nielles-Vallespin_2017_AA}
Nielles-Vallespin, S., et~al.: Assessment of myocardial microstructural
  dynamics by in vivo diffusion tensor cardiac magnetic resonance. J. Am. Coll.
  Cardiol.  69(6),  661--676 (2017)

\bibitem{Nielles-Vallespin_2013_AA}
Nielles-Vallespin, S., et~al.: In vivo diffusion tensor {MRI} of the human
  heart: reproducibility of breath-hold and navigator-based approaches. Magn.
  Reson. Med.  70(2),  454--65 (2013)

\bibitem{qin2017convolutional}
Qin, C., et~al.: Convolutional recurrent neural networks for dynamic {MR} image
  reconstruction. arXiv preprint arXiv:1712.01751  (2017)

\bibitem{ravishankar2011mr}
Ravishankar, S., Bresler, Y.: {MR} image reconstruction from highly
  undersampled k-space data by dictionary learning. IEEE Trans. on Med. Imag.
  30(5),  1028--1041 (2011)

\bibitem{ronneberger2015unet}
Ronneberger, O., et~al.: {U-Net: Convolutional Networks for Biomedical Image
  Segmentation}. MICCAI pp. 234--241 (2015)

\bibitem{schlemper2017dynamic}
Schlemper, J., et~al.: A deep cascade of convolutional neural networks for
  dynamic {MR} image reconstruction. IEEE Trans. on Med. Imag.  37(2),
  491--503 (2018)

\bibitem{schlemper2017deep}
Schlemper, J., et~al.: A deep cascade of convolutional neural networks for {MR}
  image reconstruction. In: IPMI. pp. 647--658. Springer (2017)

\bibitem{NIPS2016_6406}
Sun, J., et~al.: Deep {ADMM-Net} for compressive sensing {MRI}. In: NIPS. pp.
  10--18 (2016)

\bibitem{wu2014accelerated}
Wu, Y., et~al.: Accelerated {MR} diffusion tensor imaging using distributed
  compressed sensing. Magn. Reson. Med.  71(2),  763--772 (2014)

\bibitem{yang2017dagan}
Yang, G., et~al.: {DAGAN}: Deep de-aliasing generative adversarial networks for
  fast compressed sensing {MRI} reconstruction. IEEE Trans. on Med. Imag.
  (2018)

\end{thebibliography}




\end{document}